\documentclass[final]{cvpr}

\usepackage{times}
\usepackage{epsfig}
\usepackage{graphicx}
\usepackage{amsmath}
\usepackage{amssymb}

\usepackage{soul}

\usepackage{amsmath,amssymb,amsthm}
\usepackage{amsmath,amsthm,amsfonts,amssymb}
\usepackage{bbm}
\usepackage{mathtools}

\usepackage{url}
\usepackage{ltablex}
\usepackage{booktabs}
\usepackage{subcaption}
\usepackage{xcolor}

\usepackage[numbers]{natbib}

\usepackage{algorithm}
\usepackage{algpseudocode}
\usepackage{hyperref}

\begin{document}

\title{Voting based ensemble improves robustness of defensive models}


\author{%
  Devvrit %
  \rlap{\textsuperscript{1}},
  Minhao Cheng %
  \rlap{\textsuperscript{2}},
  Cho-Jui Hsieh %
  \rlap{\textsuperscript{2}},
  Inderjit Dhillon %
  \rlap{\textsuperscript{1}}%
}

\maketitle

\footnotetext[1]{Department of Computer Science, University of Texas at Austin, USA}
\footnotetext[2]{Department of Computer Science, University of California, Los Angeles, USA\\
All correspondence to: devvrit@cs.utexas.edu}

\begin{abstract}
Developing robust models against adversarial perturbations has been an active area of research
and many  algorithms have been proposed to train individual robust models.
Taking these pretrained robust models, we aim to study whether it is possible to create an ensemble to further improve robustness. Several previous attempts tackled this problem by ensembling the soft-label prediction and have been proved vulnerable based on the latest attack methods. 
In this paper, we show that if the robust training loss is diverse enough, a simple hard-label based voting ensemble can boost the robust error over each individual model. 
Furthermore, given a pool of robust models, we develop a principled way to select which models to ensemble. 
Finally, to verify the improved robustness, we conduct extensive experiments to study how to attack a voting-based ensemble and develop several new white-box attacks.
On CIFAR-10 dataset, 
by ensembling several state-of-the-art pre-trained defense models, 
our method can achieve a 59.8\% robust accuracy, outperforming all the existing defensive models without using additional data.
  
\end{abstract}

\section{Introduction}
Despite achieving human-level performance in many important tasks, it has been discovered that deep networks are vulnerable to adversarial perturbations---a small but human imperceptible perturbation can easily alter the prediction of a neural network~\cite{Pertub,biggio2013evasion,szegedy2013intriguing,goodfellow2014explaining}.
Since deep neural networks are being deployed in many safety-critical applications, it becomes important to develop robust defence mechanisms to make them robust against adversarial perturbations. Adversarial training \cite{goodfellow2014explaining, aleks2017deep} has become one of the key techniques to develop defenses against adversarial attacks. \cite{madry2017towards} showed that adversarial training can be formulated as solving a minimax objective function, where one conducts Projected Gradient Descent (PGD) to find an adversarial example to maximize the loss, and then update the neural network weights based on this adversarial example. After that, many variations of adversarial training has been proposed by improving the minimax objective, including TRADES~\cite{zhang2019theoretically}, MART~\cite{Wang2020Improving}, MMA~\cite{ding2018max} and many others \cite{Wang_2019_ICCV,Zheng_2020_CVPR,xiong2020improved}.
The family of adversarial training-based defense methods have become state-of-the-arts under the current strongest white-box attacks~\cite{croce2020reliable}. 

Many of these adversarially trained models, despite having different objectives, achieving similar level of robust accuracy under white-box attacks. This naturally leads to the following important question: {\bf Can we ensemble these robust models to further boost the performance? } Surprisingly, this fundamental question has not been properly answered in the literature, and previous ensemble-based methods often fail to improve the performance under white-box attacks. For example, \cite{DBLP:journals/corr/abs-1901-08846}, \cite{sen2020empir}, and \cite{NEURIPS2019_cd61a580} suggested a special loss function, different architectures, and different DNNs output representation respectively to increase the ensemble diversity but were proven non robust \cite{tramer2020adaptive}. 


In contrast to previous works that blending softmax probabilities or logit outputs of the base models (which failed to improve robustness), we consider a majority-vote ensemble mechanism in this paper. To investigate whether majority-vote ensemble can improve robust accuracy, we need to answer the following questions: 1) How can we properly evaluate the robustness of a majority-vote ensemble?  2) How can we choose a proper set of base models to achieve the best performance? 
To answer the first question, we conduct a comprehensive study on how to best attack a majority-vote ensemble, using both existing and newly developed techniques. For the second question, we show that a majority-vote ensemble can boost robust accuracy when the loss landscape of these base models are diverse enough. We verify this empirically and propose a novel framework to automatically select which base models should be included in the ensemble. 

Our contributions can be summarized below: 
\begin{itemize}
    \item 
    We show that voting-based ensemble can improve robustness over each individual robust model, if the loss is diverse enough.
    Taking a pool of $3$ recently proposed defensive models, including Trades~\cite{zhang2019theoretically}, MART~\cite{Wang2020Improving} and PGD~\cite{madry2017towards}, 
    we show that the majority-vote ensemble can achieve 57.32\% robust accuracy on CIFAR-10 against $8/255$ $\ell_\infty$ perturbation, significantly outperforming the best base model (54.86\%). 
    \item We propose a novel algorithm to select which models to ensemble given a pool of defenses. Our algorithm is able to automatically select the best 3 models among a total set of 7 state-of-the-art models to further boost robust accuracy to 59.8\% on CIFAR-10 against $8/255$ $\ell_\infty$ perturbation, achieving the state-of-the art performance without using additional data\footnote{Based on the snapshop of \url{https://github.com/fra31/auto-attack} when we finish this paper, the best publicly available model trained on CIFAR-10 achieves 56.17\% robust accuracy. Note that all the methods achieving robust accuracy beyond 59\% are using additional unlabeled data. }. 
    \item To verify that the voting based ensemble is truly robust, we carefully test existing attack methods as well as developing several novel attack algorithms to attack the ensemble. The proposed model is still robust under these adaptive attacks. 
\end{itemize}
\noindent The rest of the paper is arranged as follows. In Section \ref{related-work} we cover the related work to adversarial defense and ensemble based defense. In Section \ref{rob-ens-def} we first introduce some basic notations and explain different ensemble types we consider. We then discuss choosing between a logits-summed ensemble and a voting-based ensemble. In Section \ref{select-model-ens-0} we study when given a pool of different defense methods, how to choose which ones to use to form an ensemble. Then we introduce several novel techniques for a white-box attacks on a majority-vote ensemble. The experimental results are delivered in Section \ref{exp-eval}. 
\section{Related Work} \label{related-work}
Since the discovery of adversarial examples~\cite{szegedy2013intriguing}, many algorithms have been proposed to improve the robustness against adversarial examples \cite{madry2017towards,zhang2019theoretically,Wang2020Improving,wu2020adversarial,pang2020boosting,huang2020sat,sitawarin2020improving,Rahnama_2020_CVPR,Zheng_2020_CVPR,Xie_2019_CVPR,Qiu_2019_CVPR,Jang_2019_ICCV,Mustafa_2019_ICCV,Wang_2019_ICCV,xiong2020improved}. However, many previous techniques have shown vulnerable under stronger adaptive attacks~\cite{athalye2018obfuscated,tramer2020adaptive}. Among them, adversarial training has become one of the most reliable approaches. In this paper, we will focus on ensembling adversarially trained models.  
\paragraph{Adversarial training}
To enhance the adversarial robustness of a neural network model, the most popular method is adversarial training which iteratively uses the generated adversarial examples back into the training process. Specifically, \citet{goodfellow2014explaining} first uses adversarial examples generated by FGSM method to augment the training data, while later \citet{kurakin2016adversarial_ICLR} uses a multi-step FGSM to further improve adversarial robustness. \citet{madry2017towards} formalize this iterative data augmentation process into a min-max optimization problem and propose to use PGD attack (similar to multi-step FGSM) to find adversarial examples for each batch. It shows the adversarial trained model could achieve a relatively good adversarial robustness even facing with very strong attacks \citep{athalye2018obfuscated}. Based on the min-max framework, \citet{zhang2019theoretically} proposes TRADES, a theoretical based framework to adjust the trade-off between adversarial robustness and generalization. \citet{Wang2020Improving} later introduce label correctness into the TRADES and propose MART to improve the overall performance. Recently, \citet{schmidt2018adversarially} finds the sample complexity of robust learning can be significantly larger than that of standard learning. Since then, several works \citep{carmon2019unlabeled,alayrac2019labels} have been introduced to use unsupervised or semi-supervised method to introduce more data into training process and shows it could further push the limit on the robustness. Many other recent defense methods also follow this min-max optimization framework \cite{Wang_2019_ICCV,Zheng_2020_CVPR,xiong2020improved,pang2020boosting}

\paragraph{Previous research on ensemble-based defense}
There has been some work on using ensemble to boost the robustness of DNNs but most of them have been shown to fail under different attacks. \citet{liu2018towards} added random noise to form an ensemble and showed that it could achieve a better robustness. \citet{DBLP:journals/corr/HeWCCS17} considered several ensemble methods using weak individual models and showed them non-robust. \citet{tramr2017ensemble} used ensemble adversarial training by collecting adversarial examples from different models to train a single robust model. Elaborate black-box attack which enhanced transferability showed \cite{tramr2017ensemble} non-robust though. \citet{strauss2017ensemble} studied popular ensemble methods like bagging, adding gaussian noise during training, and using different training architecture. But their analysis are under weak attacks and they don't consider majority-vote ensembles.
\citet{liu2018towards} proposed random self-ensemble for defense, but it is based on ensembling a set of randomized models with different random seed, and their method cannot be used to blend several different models. 
\citet{DBLP:journals/corr/abs-1901-08846} suggested training an ensemble using a specially crafted loss function to increase the robustness. \citet{sen2020empir} use architectures with different precisions to form a majority-vote ensemble. \citet{NEURIPS2019_cd61a580} aimed at introducing enough diversity in models forming an ensemble and use ideas from error correcting code for robust classification. However, these all three ensemble defenses were shown in \citet{tramer2020adaptive} to be ineffective under several attacks. 

\section{Robustness of Ensemble Defense} \label{rob-ens-def}
In this section, we will first introduce the proposed majority-vote ensemble and provide empirical evidences showing it outperforms sum-of-logit based ensembles. Then we discuss how to select the right subset of models to ensemble in Section~\ref{select-model-ens-0}. Finally, we discuss existing and newly proposed ways to attack majority-vote ensemble in Section~\ref{attacks}. 

\subsection{Ensemble methods}

We consider a $C$-way classification problem and assume there are $n$ neural network models that we want to ensemble. 
We denote an input as $(\mathbf{x}, y)$ where $\mathbf{x}\in \mathbb{R}^{d}$ is the input image and $y\in[C]$ is the class label.
The logits of $i^{th}$ neural network is defined by $f_i(\mathbf{x})\in \mathbb{R}^{C}$, $i\in [n]$. The inputs to the softmax function
are called logits. The label assigned by each network is denoted by 
\begin{equation*}
    F_i(\mathbf{x}) := \arg\max_{c=1, \dots, C} f_i(\mathbf{x})_c.
\end{equation*}

\paragraph{Logits-summed ensemble} \label{logits-summed-ensemble}
Given $n$ networks in the ensemble, most of the previous methods use the \textit{logits-summed} ensemble \cite{DBLP:journals/corr/abs-1901-08846,strauss2017ensemble}, where the decision function $f(\mathbf{x})$ is defined as 
\begin{align*}
    f(\mathbf{x}) := \sum\nolimits_{i=1}^{n}f_i(\mathbf{x}). 
\end{align*}
Informally, the logit output by the ensemble is  the sum of logits by individual networks comprising the ensemble.


\begin{table*}[h!]
\caption{Average $\cos(\theta)$ where $\theta$ is the angle between adversarial perturbation on pair of models under the same attack. Left (a) is on PGD-CE attack, right (b) is on C\&W attack. We take TRADES \cite{zhang2019theoretically}, MART \cite{Wang2020Improving}, TRADES-AWP \cite{wu2020adversarial}, PGD \cite{madry2017towards}, and PGD+HE \cite{pang2020boosting} and represent them as $0,1,2,3,4$ respectively, for brevity} \label{table0}
\centering

\begin{subtable}{.5\linewidth}
\caption{}
\centering

\begin{tabular}{|p{0.13\linewidth}|p{0.08\linewidth}|p{0.08\linewidth}|p{0.08\linewidth}|p{0.08\linewidth}|p{0.08\linewidth}|}
\hline
\textit{Defense} & \textit{0} & \textit{1} &\textit{2} &\textit{3} &\textit{4} \\
\hline
\textit{0} & \textbf{0.485} & 0.365 & 0.435 & 0.2 & 0.265\\
\textit{1} & 0.365 & \textbf{0.477} & 0.36 & 0.203 & 0.328\\
\textit{2} & 0.435 & 0.36 & \textbf{0.51} & 0.21 & 0.27\\
\textit{3} & 0.2 & 0.203 & 0.21 & \textbf{0.24} & 0.203\\
\textit{4} & 0.265 & 0.328 & 0.27 & 0.203 & \textbf{0.49}\\
\hline
\end{tabular}
\end{subtable}%
\begin{subtable}{.5\linewidth}
\caption{}
\centering

\begin{tabular}{|p{0.13\linewidth}|p{0.08\linewidth}|p{0.08\linewidth}|p{0.08\linewidth}|p{0.08\linewidth}|p{0.08\linewidth}|}
\hline
\textit{Defense} & \textit{0} & \textit{1} &\textit{2} &\textit{3} &\textit{4} \\
\hline
\textit{0} & \textbf{0.39} & 0.287 & 0.343 & 0.16 & 0.276\\
\textit{1} & 0.287 & \textbf{0.395} & 0.28 & 0.179 & 0.313\\
\textit{2} & 0.343 & 0.28 & \textbf{0.41} & 0.167 & 0.26\\
\textit{3} & 0.16 & 0.179 & 0.167 & \textbf{0.197} & 0.176\\
\textit{4} & 0.276 & 0.313 & 0.26 & 0.176 & \textbf{0.408}\\
\hline
\end{tabular}
\end{subtable}

\end{table*}

\paragraph{Majority-vote ensemble} \label{majority-vote-ensemble}
Given $n$ networks in the ensemble, we define the \textit{majority-vote} ensemble as a network with output $F(x) \in \mathbb{R}^{C}$ as
\begin{equation*}
    F(\mathbf{x}) := \arg\max_{c=1, \dots, C} \sum\nolimits_{i=1}^n F_i(\mathbf{x}). 
\end{equation*}
If more than one components have max value then we assign $1$ to any one of them arbitrarily.
Informally, the majority-vote ensemble outputs a one-hot vector with prediction as the class having maximum vote by comprising individual networks in the ensemble.

\subsection{Shifting from logits-summed ensemble to majority-voting ensemble:}
We started by asking ourselves the natural first question: \textbf{Is it sufficient to just take few (differently initialized) models of the same defense type and make a logits-summed ensemble?} 
We conduct experiments for ensembling 3 models and the robustness improvement was only marginal. For example, against one of the most effective adversarial attack, Autoattack \cite{croce2020reliable},  
TRADES \cite{zhang2019theoretically} robust accuracy improved from $54.08\%$ of an individual TRADES model to $56.4\%$ of the logits-summed ensemble. For MART \cite{Wang2020Improving}, the improvement was less 
(from $54.86\%$ to $55.7\%$). 

The natural question we asked next was: \textbf{Can we take different defense models and ensemble them rather?} We formed a logits-summed ensemble of MART, TRADES, and PGD \cite{madry2017towards} model and found the robust accuracy $51.9\%$, which is lesser than before. Upon inspection, we found that most of the points were able to only fool just one of the three models. The individual robust accuracy on each of the TRADES, MART, and PGD model against the adversarial points generated for their logits-summed ensemble was 65.5\%, 66.8\%, and 48.8\% respectively. This means that PGD model logits were driving the sum of the logits towards the wrong class. As two of the three models were still predicting correct for most of the examples, we rather decide to use majority-vote ensemble. Using the later developed (Section \ref{attacks}) white-box attacks for voting-based ensemble, the worst accuracy we got for TRDAES, MART, PGD voting-based ensemble is $57.4\%$ --- much higher than $51.9\%$ when using logits-summed ensemble.

This led to the following intuition: \textbf{Do models trained with different loss functions have diverse decision boundaries?} At first, this looks reasonable because loss functions have a direct implication on the geometry of decision boundary. We took pairs of models, attacked them individually with the same attack, and measured the average cosine of angle between perturbations generated for the pair of models. Let $\theta$ represent this angle for a given point $\mathbf{x}$. We'd expect that if indeed models trained on different loss functions are more diverse, their adversarial perturbations will be more diverse too (hence smaller average $\cos(\theta)$) compared to a pair of models which are trained on the same loss function (hence larger average $\cos(\theta)$). 
We took $5$ recently proposed state-of-the art defenses\footnote{we would have taken more but could find training code available for only 5 of them} (mentioned in Table \ref{table0}). We trained $2$ (differently initialized) models of each defense method, attacked them using PGD attack \cite{madry2017towards} and C\&W attack \cite{Carlini_2017}, and measured average (taken over points) $\cos(\theta)$ of the perturbations for all $5*5=25$ pairs of models. We report the numbers in Table \ref{table0} and notice the results in alignment to our intuition. We believe this is some metric, if not the perfect metric, to show the difference in diverseness of decision boundary for pairs of models trained on the same loss function compared to different loss functions.

Another question that could be asked is: \textbf{What if we rather use a pool of similarly-robust defenses?} In this case, one could expect the logits-summed ensemble to perform well. In order to answer this, we considered the following $5$ recently proposed state-of-the-art defenses which have robust accuracies within $0.7\%$ of each other (Table \ref{table4}): TRADES \cite{zhang2019theoretically}, MART \cite{Wang2020Improving}, TRADES+SAT \cite{huang2020selfadaptive}, ATES \cite{sitawarin2020improving}, and PGD+HE \cite{pang2020boosting}. We formed $3$ network ensembles, hence $5^3=125$ possibilities (remember all models in an ensemble are differently randomly initialized). Testing these many ensembles is tough, hence we tested a big and good representative subset of them. This includes all possible ${5\choose3}=10$ ensembles consisting three models all of different defense methods, $7$ ensembles consisting $2$ models of one defense type and $1$ of another defense type, and $3$ ensembles consisting all three models of same defense type. The highest accuracy against Autoattack we got was $57.9\%$. However, just considering the ${5\choose3}=10$ possible \textbf{majority-vote} ensembles consisting three models all of different defense methods, we got $59.17\%$ robust accuracy. The majority-vote ensemble accuracy is measured using WA-autoattack, the strongest white-box attack we found as discussed later in Sections \ref{attacks}, \ref{strong-attacks}.

In order to take advantage of diverse decision boundaries, majority-vote ensembles seem to be the natural fit compared to a logits-summed ensemble. It has been widely observed that DNNs predict the wrong class with high confidence against an adversarial example \cite{szegedy2014intriguing,moosavidezfooli2016deepfool}. Hence, the sum of logits could be driven to wrong class due to a single model being non-robust on a given point, even if the other models in the ensemble may be predicting correct (as we saw in PGD, MART, TRADES logits-summed ensemble). All the above observations motivate us to investigate majority-vote ensemble over a logits-summed ensemble.



\subsection{Model Selection for majority-vote ensemble} \label{select-model-ens-0}
Given $n$ individual defenses, our aim is to select $k$ defenses out of the $n$ available to form a majority-vote ensemble. This problem isn't easy to solve as it requires selecting models based on how individually robust they are, as well as how diverse they are with respect to each other. In the following, we model this problem from an optimization point of view and derive an algorithm for model selection. 
Let $g_i(\cdot)$, where $i\in \left[{n\choose k}\right]$ and $g_i(\cdot)\in \mathbbm{R}^{C}$ denote the $i^{th}$ ensemble formed of $k$ models. We assume data follow the distribution $(\mathbf{x},y)\sim D$ and use $\mathcal{B}$ to denote the allowed perturbation set. 
For example, a typical $\ell_p$ norm threat model assumes $\mathcal{B} = \{\boldsymbol{\delta} \mid \|\boldsymbol{\delta}\|_p\leq \epsilon\}$, where $\epsilon$ is preset threshold. So the adversary is allowed to perturb an original point $\mathbf{x}$ to any point $\mathbf{x}'$ such that $\mathbf{x}'-\mathbf{x}\in \mathcal{B}$. 

Let $sc(g_i, \mathcal{B}, D)$ denote robust accuracy of $i^{th}$ ensemble defined as
\begin{equation*}
sc(g_i, \mathcal{B}, D) = \underset{(\mathbf{x},y)\sim D}{\mathbb{E}}\min_{\boldsymbol{\delta}\in \mathcal{B}}[\mathbbm{1}_{g_i(\mathbf{x}+\boldsymbol{\delta})=y}],
\end{equation*}
where the inner minimization corresponds to the adversary that finds the worst-case perturbation within set $\mathcal{B}$ for each input, and $y$ is the correct label. 
Finding the best ensemble can then be formally written as finding the best $i$ from all the ${n\choose k}$ cases to maximize robust accuracy: 
\begin{equation}
    \max_i sc(g_i, \mathcal{B}, D).
    \label{eq:choosing-ensemble}
\end{equation}
However, solving \eqref{eq:choosing-ensemble} is intractable
because to (approximately) solving the inner minimization, a strong adversarial attack is required to be conducted for all the ${n\choose k}$ ensembles, and a large-enough subset of samples needs to be tested in order to obtain a good estimation of robust accuracy. Even if we use a simple PGD-based attack, it is computational infeasible to run it on all the $O({n\choose k})$ cases. 

Hence, we propose a novel algorithm to approximately solve \eqref{eq:choosing-ensemble} in tractable time. The main idea is to use the adversarial examples generated by base models instead of ensemble models to evaluate robust accuracy. 
We take $r$ random test/validation points $\{\mathbf{x_1},\ldots,\mathbf{x_r}\}$ with true labels $\{y_1,\ldots,y_r\}$ where $y_i\in [C]$. We generate perturbation for each point by attacking any one of the $n$ models. To maintain uniformity, we use the first model to generate perturbations for the first $r/n$ points, the second model for next $r/n$ points, and so on. 
Let $\{\mathbf{x_1}',\ldots,\mathbf{x_r}'\}$ denote these $r$ adversarial examples. For each ${n \choose k}$ possible ensembles, we calculate its score as the number of adversarial points predicted corrected by at least $\lceil k/2 \rceil$ out of the $k$ models in the ensemble.
This is because we're focusing on majority-vote ensemble and so using its definition to calculate the ensemble's score. Finally, we choose the ensemble with the highest score. The algorithm is also presented in appendix \ref{appendix-select-model-ens}

The approximate evaluation scheme introduced above can reduce the number of attacks from $r\times {n\choose k }$ to $r\times n$, which makes model selection feasible. In our current implementation, there are only few models to ensemble so we will evaluate the approximate score for each of ${n \choose k}$ models and select the best. If $n$ goes larger, we can further use a genetic algorithm to conduct model selection based on this approximate scoring function.  

Intuitively, testing a model on adversarial points captures the model's robustness. And testing the adversarial points generated by one model on another tests transferability and hence the diverseness of different models with respect to each other. We later show this heuristic gives a reasonable approximation to the actual eq.\eqref{eq:choosing-ensemble} objective in Section \ref{select-model-ens}, by considering a pool of $7$ different defenses.

\subsection{How to attack a majority-vote ensemble?} \label{attacks}
As we want to objectively test the robustness, we perform white-box attacks on all the ensembles. 
Since the voting mechanism is a {\bf discrete} process, the gradient of ensemble does not exist and it is nontrivial to attack a voting based ensemble.
Currently the only method used in the literature for attacking voting-based ensemble is to approximate it by a logits-summed ensmeble. This method, called {\bf logits-summed attack} in our paper, has been used in \cite{tramer2020adaptive} to break the defense in  \cite{sen2020empir} and was also mentioned in~\cite{athalye2018obfuscated}. 
On the other hand, although BPDA attack~\cite{athalye2018obfuscated} can handle some discrete models, they only cover  discrete processes that can be approximated by an identity mapping and applied a straight-through estimator for attack, which is not applicable for majority-vote ensemble. 

There hasn't been \textbf{any} systematic study on developing white-box attacks to test majority-vote ensemble.
Here, in addition to the logits-summed attack, we propose three other attack techniques which could be used on top of any existing white-box attacks to test majority-vote ensemble.


\subsubsection{Weakest-attacked attack} \label{WA}
Let the point in consideration be $(\mathbf{x},y)$. Let there be $n$ networks in the ensemble. The attacker calculates logits $f_i(\mathbf{x})$ for all $n$ networks. If the majority vote $F(x)$ for majority-vote ensemble (as defined in Section \ref{majority-vote-ensemble}) is already the non-true class, then the attacker is done. Otherwise, among the individual models (also referred to networks in this work) which predict correct label $y$, the attacker chooses the weakest model and performs a local small step attack on it. The weakest model among those which predict the true class $y$ is determined by calculating the probability of individual models on true class $y$ and choose the one with lowest probability. Let $\mathbf{y} \in \mathbb{R}^{C}$ represent the one-hot vector with $1$ at component $y$ and 0 otherwise. Mathematically the weakest model output can be represented as $f_{k}(\mathbf{x})$ where
\begin{align*}
    k &= \arg\min_{i\in [n]}
    (\mathbbm{1}_{F_i(\mathbf{x})=\mathbf{y}}\cdot (p_i(\mathbf{x}))_{y} + \mathbbm{1}_{F_i(\mathbf{x})\neq \mathbf{y}}\cdot 1), \\
    &\text{where } p_i(\mathbf{x}) = Softmax(f_i(\mathbf{x}))
\end{align*}
and $\mathbbm{1}$ is an indicator variable and $(p_i(\mathbf{x}))_{y}$ refer to the $y^{th}$ component of vector $p_i(\mathbf{x})$.

Take an example of an ensemble consisting of three models $\{m_1,m_2,m_3\}$. Let the clean data point be $x$ and its true class be $y$. The attacker checks if the majority vote is already a non-true class. If so, the attacker is done. Let's say this is not the case and $\{m_1,m_2\}$ predict the true class $y$. Also let the probability of class $y$ predicted by $m_1$ be less than that predicted by $m_2$. Most of the existing white-box based attacks happen by taking multiple small steps towards final perturbation. In this case, the attacker performs a local attack on $m_1$ and moves $x$ to $x+\epsilon$. The attacker again passes this $x+\epsilon$ through all three models $\{m_1,m_2,m_3\}$. Let's say still $\{m_1,m_2\}$ predict the true class $y$ but this time probability of class $y$ by $m_2$ is less than that of $m_1$. In this step, the attacker performs a local attack on $m_2$ to to get $x+\epsilon+\epsilon'$. This is repeated until either the number of steps are over or the majority vote becomes a non-true class. Intuitively, the attacker performs a greedy attack over the three models to turn the majority vote to a non-true class. We present the attack formally in appendix \ref{appendix-WA}.

Note that similar to logits-summed attack technique, this attack technique could also use any existing white-box attack. We never fixed what attack is applied on the locally weakest model. In this work we consider PGD with CE loss attack \cite{aleks2017deep}, FAB-attack \cite{croce2019minimally}, autoattack \cite{croce2020reliable}, and $l_{\infty}$ B\&B attack \cite{brendel2019accurate} with this attack technique.

\subsubsection{Objective-summed attack} \label{CW}
Most of the existing white-box attacks decrease/increase a particular objective. For example, PGD attack with CE loss function maximizes the CE loss, C\&W attack minimizes the C\&W loss \cite{Carlini_2017}. Therefore another attack technique is to take sum of this attack objective values on individual models of the ensemble and decrease/increase it. Notice that just like previous techniques, even this method can work with any white-box attack. In this work, we particular consider C\&W loss objective. That is, we sum the C\&W loss over all the models comprising the ensemble, and decrease it using PGD.

\subsubsection{majority-attack} \label{majority-attack}
Rather than using the previous three introduced attack technique over the entire ensemble, one can use them over just a subset of $\lceil n/2 \rceil$ models among the $n$ models comprising the ensemble. Intuitively, as majority-vote ensemble needs to misclassify any $\lceil n/2 \rceil$ models, the attacker focuses on doing just that. One needs to consider all possible ${n \choose n/2}$ combination though, hence this attack technique isn't very efficient. But for small $n$ one can use it. In this work, we use Objective-summed attack technique (Section \ref{CW}) with C\&W attack on an ensemble consisting $3$ models and attack all possible ${3 \choose 2}$ pairs of models of the ensemble.

\section{Experimental Evaluations} \label{exp-eval}
In this section, we select various existing defenses and perform extensive experiment on them. We first introduce certain notations, then ensemble $3$ (hand-picked) defenses and perform a more thorough analysis on them. We then use the algorithm mentioned in Section \ref{select-model-ens-0} on the entire pool of $7$ models, form various ensembles, and test them under strong attacks.


\subsection{Ensemble models} \label{ensemble-models}
We select a pool of $7$ state-of-the-art defenses: PGD \cite{madry2017towards}, TRADES \cite{zhang2019theoretically}, MART \cite{Wang2020Improving}, TRADES-AWP \cite{wu2020adversarial}, PGD+HE \cite{pang2020boosting}, TRADES+SAT \cite{huang2020sat}, and ATES \cite{sitawarin2020improving}. We pick $3$ models to form the ensemble, each model being one of the above $7$ defense types. Every individual model in all the ensembles is differently initialized when training. 
The final ensemble output is the majority vote of individual models. 
If there’s no majority vote, the ensemble outputs one of the model predictions randomly.

We report as a baseline the individual model accuracy against PGD attack with CE loss, FAB attack, and C\&W attack in Table \ref{table4} (a). The parameters for these attacks are mentioned in Section \ref{attacks-1}. We later in Sections \ref{strong-attacks}, \ref{select-model-ens} test a few ensembles against stronger attacks. Specifically, in Table \ref{table4} (b) we test the above $7$ models accuracy against these stronger attacks. We trained TRADES, MART, and PGD model from scratch hence they might be slightly different in performance compared to the ones available online.

\paragraph{Notations} \label{notations}
In tables and a few other places, for brevity, we represent TRADES model with the letter T, MART model with letter M, and PGD trained model with letter P. T-r/M-r represents a TRADES/MART model trained with $\beta=r$. TRADES and MART loss function is based on a mix of clean accuracy and robustness term, where robustness is controlled by the parameter $\beta$. Unless otherwise specified, T=T-6 and M=M-6. Logits-summed-PGD-CE attack is represented by LS-PGD, Logits-summed-FAB is represented by LS-FAB, Weakest-attacked-PGD-CE is represented by WA-PGD, and Weakest-attacked-FAB is represented by WA-FAB. Every experiment on an ensemble is repeated $2-4$ times. We report the mean values in the table.

\subsection{Attacks} \label{attacks-1}
We use the following $5$ white-box attacks: 1) logits-summed-CE (LS-CE): PGD attack with CE loss function using logits-summed attack technique (Section \ref{attacks}, \cite{tramer2020adaptive}); 2) logits-summed-FAB (LS-FAB): FAB attack using logits-summed attack technique (Section \ref{attacks}, \cite{tramer2020adaptive}); 3) weakest-attacked-CE (WA-CE): PGD attack with CE loss using the technique as explained in Section \ref{WA}; 4) weakest-attacked-FAB (WA-FAB): FAB attack using the technique in Section \ref{WA}; 5) C\&W attack using objective-summed attack technique, as mentioned in Section \ref{CW}.

We consider $l_{\infty}$ attack with $\epsilon=0.03$. For PGD-CE attack, we set the number of iterations as 150 and the learning rate to be $0.007$. For FAB attack \cite{croce2019minimally} we run for 25 iterations and keep rest of the parameters default from advertorch implementation\footnote{https://github.com/BorealisAI/advertorch} \cite{ding2019advertorch}. For C\&W attack \cite{Carlini_2017}, we set the number of iterations to be $150$, learning rate $0.007$, and $\kappa=0$.


\begin{table*}[h!]
\caption{Mean accuracy of individual models. Left (a) attacks have parameters as mentioned in Section \ref{attacks-1}, while right (b) are stronger attacks with parameters mentioned in Section \ref{strong-attacks}} \label{table4}
\centering

\begin{subtable}{.5\linewidth}
\caption{}
\centering
\begin{tabular}{|p{0.15\linewidth}|p{0.16\linewidth}|p{0.13\linewidth}|p{0.13\linewidth}|}
\hline
\textit{Ensemble} & \textit{PGD-CE}
 &\textit{FAB} & \textit{C\&W}\\
 \hline
  T (T-6) &  55.7\%  & 53.97\% & 54.08\% \\
  T-4 &  54.5\%  & 53.3\% & 53.65\% \\
  T-2 &  52.9\%  & 52.09\% & 52.3\%\\
  T-1 &  50.28\%  & 50.06\% & 50.5\% \\
  M (M-6) &  57.9\%  & 54.13\% & 54.86\% \\
  M-4 &  57.99\%  & 54.31\% & 54.8\% \\
  M-2 &  57.17\%  & 53.8\% & 54.54\% \\
  M-1 &  56.42\%  & 53.38\% & 54.56\% \\
  P &  49.1\%  & 50.2\% & 49.2\% \\
\hline
\end{tabular}
\end{subtable}%
\begin{subtable}{.5\linewidth}
\caption{}
\centering
\begin{tabular}{|p{0.35\linewidth}|p{0.2\linewidth}|p{0.17\linewidth}|}
\hline
\textit{Ensemble} & \textit{Strong-FAB}
 &\textit{autoattack}\\
 \hline
  PGD \cite{madry2017towards} & 48.72\% & 46.94\%\\
  TRADES \cite{zhang2019theoretically} & 53.97\% & 54.08\% \\
  MART \cite{Wang2020Improving} & 54.13\% & 54.86\% \\
  ATES \cite{sitawarin2020improving} & 53.56\% & 53.93\% \\
  TRADES-AWP \cite{wu2020adversarial} & 58.09\% & 57.6\% \\
  TRADES+SAT \cite{huang2020selfadaptive} & 55.27\% & 54.65\% \\
  PGD+HE \cite{pang2020boosting} & 55.24\% & 54.59\% \\
\hline
\end{tabular}
\end{subtable}
\end{table*}

\begin{table*}[h!]
\caption{Ensemble robust accuracy of Trades and Mart trained with diff $\beta$} \label{table1}
\centering

\begin{tabular}{|p{0.12\linewidth}|p{0.09\linewidth}|p{0.09\linewidth}|p{0.09\linewidth}|p{0.09\linewidth}|p{0.09\linewidth}|p{0.22\linewidth}|}
\hline
\textit{Ensemble} & \textit{LS-PGD} & \textit{LS-FAB} &\textit{C\&W} &\textit{WA-PGD} &\textit{WA-FAB} &\textit{acc. against best attack}\\
\hline
  T-6,T-6,T-4 &  56.89\%  & 69.86\% & 58.16\%   & 58.99\%  & 57.36\%  & 56.89\%  \\
  T-6,T-6,T-2 &  57.17\%  & 70.87\% & 58.35\%   & 59.13\%  & 57.52\%  & 57.17\%  \\
  T-6,T-6,T-1 &  57.55\%  & 71.1\% & 58.96\%   & 59.5\%  & 57.87\%  & 57.55\%  \\
  M-6,M-6,M-4 &  57.38\%  & 69.36\% & 58.2\%  & 61.01\%  & 56.94\%  & 56.94\%  \\
  M-6,M-6,M-2 &  57.5\%  & 69.05\% & 58.92\%   & 61.5\%  & 57.4\%  & 57.4\%  \\
M-6,M-6,M-1 &  57.73\%  & 69.72\% & 58.55\%   & 61.7\%  & 57.55\%  & 57.55\%  \\
\hline
\end{tabular}
\end{table*}

\begin{table*}[h!]
\caption{Robust accuracy of ensembles of different models. Note that T,T,T indicate an ensemble of three Trades models trained with different random initializations. 
} \label{table2}
\centering

\begin{tabular}{|p{0.12\linewidth}|p{0.09\linewidth}|p{0.09\linewidth}|p{0.09\linewidth}|p{0.09\linewidth}|p{0.09\linewidth}|p{0.22\linewidth}|}
\hline
\textit{Ensemble} & \textit{LS-PGD} & \textit{LS-FAB} &\textit{C\&W} &\textit{WA-PGD} &\textit{WA-FAB} &\textit{acc. against best attack}\\
\hline
  T,T,T &  57.18\%  & 70.1\% & 58.6\%  & 58.88\%  & 57.14\%  & 57.14\%  \\
  M,M,M &  57.4\%  & 69.17\% & 57.77\%  & 60.93\%  & 57\%  & 57\%  \\
  P,P,P &  55.1\%  & 70.37\% & 56.56\% & 54.15\%  & 55.67\%  & 54.15\%  \\
  T,T,M &  57.74\%  & 70.3\% & 59.33\%  & 61.78\%  & 58.39\%  & 57.74\%  \\
  T,T,P &  62.82\%  & 76.92\% & 60.46\%  & 59.8\%  & 58.2\%  & 58.2\%  \\
  M,M,T &  58.22\%  & 69.69\% & 59.17\%  & 61.44\%  & 57.47\%  & 57.47\%  \\
  M,M,P &  65.28\%  & 76.79\% & 61\%  & 61.79\%  & 57.69\%  & 57.69\%  \\
  P,M,T &  63.18\%  & 77.05\% & 61.26\%  & 63.29\%  & 59.63\%  & 59.63\%  \\
\hline
\end{tabular}
\end{table*}

\begin{table*}[h!]
\caption{Robust accuracy of selected ensembles against stronger attacks} \label{table3}
\centering

\begin{tabular}{|p{0.09\linewidth}|p{0.15\linewidth}|p{0.06\linewidth}|p{0.09\linewidth}|p{0.12\linewidth}|p{0.13\linewidth}|p{0.15\linewidth}|}
\hline
\textit{Ensemble} &\textit{Strong WA-FAB} &\textit{RayS} &\textit{WA-B\&B} &\textit{C\&W over 2} &\textit{LS-autoattack} &\textit{WA-autoattack} \\
\hline
 T,T,T &  56.62\%  & 62.6\% & 56.9\% & 58.28\% & 58.85\% & 54.9\% \\
 T,T,P &  57.54\%  & 61.6\% & 57.56\% & 58.83\% & 64.95\% & 56.03\%  \\
 P,M,T &  59.03\%  & 62.57\% & 58.68\% & 60.59\% & 65.22\% & 57.32\% \\
\hline

\end{tabular}

\end{table*}

\subsection{Ensemble of models trained with same objective function but different hyper-parameters}

Before testing models with different training objectives, we first conduct experiments to investigate whether ensembling models trained with varied hyper-parameters on the same objective function can improve robustness. 
We form 3 ensembles all forming by three TRADES models but trained with different $\beta$ parameter. Among the three individual models in each of these $3$ ensembles, two of the models are fixed as T-6 models, while the third one is kept either of T-4, T-2, and T-1. Though individual Trades model accuracy goes down with $\beta$, we notice a consistent improvement in accuracy of the ensemble (Table \ref{table1}). We believe this is because as $\beta$ takes lower values, the overall loss functions becomes more and more different than the one with $\beta=6$. Hence, even though the individual model accuracy is decreasing, the ensemble diversity increases by small margins leading to an overall better ensemble defense. A similar trend is observed across most of the attacks by using MART with different $\beta$(s). That is, two of the models in the ensemble is fixed at $\beta=6$ while the third model is chosen among $\beta=4/2/1$ (Table \ref{table1}). We see later in Table \ref{table2} that ensembles [T,T,P] and [M,M,P] have a more pronounced increase because this third model is trained on a completely different loss function like the \cite{aleks2017deep} min-max loss function with CE loss.

\subsection{Ensemble of models trained with different objective functions}
Next we hand-pick some sets of base models trained with different objective functions to demonstrate a significant boost of robust accuracy when ensemble a set of diverse models. 
Table \ref{table2} show accuracy of ensembles having three similar models, two similar and one different model, and all three different models (last row). The last column denotes the worst accuracy among the $5$ attacks. We notice a consistent improvement in accuracy of any ensemble comprising three similar model to when replaced by one different model and the rest remaining same, to all three different models, on all the attacks.

\paragraph{Stronger attacks to validate robust accuracy} \label{strong-attacks}
We further test a selected representative 3 ensembles - [T,T,T], [T,T,P], and [P,M,T] against stronger attack to validate robust accuracy and the accuracy trend. We take these particular ensembles for the following reasons: 1) TRADES is a strong defense hence forms a good baseline of using three similar model ensemble; 2) Replacing one of the TRADES by MART would not support our claim as strongly as replacing it with PGD. This is because PGD is substantially less robust (refer Table \ref{table4}) compared to TRADES, being almost $5-6\%$ less robust across various attacks. While MART is still almost similar robust to TRADES across various attacks (Table \ref{table4}). Hence, still observing an improvement in the ensemble of [T,T,P] compared to [T,T,T] supports our hypothesis strongly; 3) [P,M,T] consists of all three individual models trained on different loss functions, inducing the highest diversity.


We test the above ensembles against 1) a stronger WA-FAB attack with default parameters from advertorch\footnote{https://github.com/BorealisAI/advertorch} 2) in order to show that there's no gradient masking involved we test against RayS \cite{chen2020rays}, a blackbox attack, on $\sim 2500$ random samples; 3) WA-B\&B attack: L-inf B\&B attack \cite{brendel2019accurate} on the locally weakest model technique as mentioned in Section \ref{WA}. We use the foolbox implementation\footnote{https://foolbox.jonasrauber.de/} \cite{rauber2017foolboxnative, rauber2017foolbox} with default parameters, making it a strong attack; 4) C\&W attack as introduced in Section \ref{majority-attack}. We report the worst accuracy obtained among the possible ${3 \choose 2}$ subsets over which C\&W loss objective is summed and optimized. We refer to this attack as "C\&W over 2" in Table \ref{table3}; 5) LS-autoattack: Autoattack \cite{croce2020reliable} using logits-summed attack technique (Section \ref{attacks}); 6) WA-autoattack: Autoattack using the weakest-attacked technique as mentioned in Section \ref{WA}.

Autoattack \cite{croce2020reliable} introduces an ensemble of different attacks including an improved version of PGD (APGD or apgd-ce for improved PGD-CE), a targeted PGD attack with a new loss function (apgd-t), FAB, and square attack \cite{andriushchenko2020square}. The authors suggest to use 'standard' attack with an ensemble of ['apgd-ce', 'apgd-t', 'FAB', 'square'] attacks. But we noticed using auto attack with these constituent attacks was quite slow to evaluate the entire testset. Furthermore, taking a random sample of $1000$ points we noticed that ['FAB', 'square'] could introduce only $1$ extra successful adversarial example compared to what ['apgd-ce', 'apgd-t'] could generate. Hence, we use a custom autoattack using ensemble of ['apgd-ce', 'apgd-t'] with better params than standard ensemble attack - apgd.n\_restarts = 2, apgd\_targeted.n\_restarts = 2, apgd\_targeted.n\_target\_classes = 9, while other parameters being same as default\footnote{https://github.com/fra31/auto-attack}

We report the ensemble accuracy against above mentioned strong attacks in Table \ref{table3}. We still observe the same trend consistent across all the attacks supporting our claim that individual models trained on different loss function have diverse decision boundary thus possibly leading to more diversity when formed an ensemble. Also notice that WA-autoattack and WA-FAB are two of the strongest attacks, which we'll use to test more ensembles in the next subsection.

\begin{table}[h!]
\caption{Accuracy of selected ensembles} \label{table-final}
\begin{tabular}{|p{0.2\linewidth}|p{0.3\linewidth}|p{0.3\linewidth}|}
\hline
\textit{Ensemble} & \textit{Strong WA-FAB}
 &\textit{WA-autoattack} \\
 \hline
  P,TA,PH & 61.57\% & 59.84\% \\
  TA,TS,PH & 61.42\% & 59.78\% \\
  P,A,TA & 60.86\% & 59.42\% \\
  P,TS,PH & 60.97\% & 59.2\% \\
  A,TS,PH & 60.89\% & 59.17\% \\
\hline
\end{tabular}
\end{table}

\subsection{Selecting individual models from a pool to form an ensemble} \label{select-model-ens}
We test the algorithm described in Section \ref{select-model-ens-0} to select individual models that forms the strongest ensemble. We show empirically that this heuristic works with reasonable approximation. As mentioned, we choose $n=7$ different defense models: PGD \cite{madry2017towards}, TRADES \cite{zhang2019theoretically}, MART \cite{Wang2020Improving}, Trades-AWP \cite{wu2020adversarial}, PGD+HE \cite{pang2020boosting}, TRADES+SAT \cite{huang2020sat}, and ATES \cite{sitawarin2020improving}. We fix
$k=3$, and choose $r \approx 4500$ random points. We ran the WA-autoattack and WA-FAB attack on all ${7\choose 3}=35$ ensembles. We measure the effectiveness of the algorithm using Kendell's $\tau$ statistic of predicted score by the algorithm and the actual accuracy of the ensembles. Kendel's $\tau$ is used to measure ordinal association between two quantities. The kendell's $\tau$ average value over three runs of the algorithm is $+0.53$ and $+0.503$ for WA-autoattack and WA-FAB attack respectively. 
Note that the range of kendell's $\tau$ is $[-1, +1]$, where $0$ indicates zero correlation, and $>0.5$ typically indicates a moderate  positive correlation. 
Moreover, {\bf in each run we got the highest accuracy ensemble within the top 3 ranked/scored by our algorithm}. We provide the supporting figures in appendix \ref{appendix-select-model-ens}.
Finally, we report accuracy on few of the best ensembles among these ${7\choose 3}$ ensembles in Table \ref{table-final} against WA-autoattack and WA-FAB attack. We represent PGD by P, MART by M, TRADES by T, TRADES-AWP by TA, PGD+HE by PH, TRADES+SAT by TS, and ATES by A for brevity in Table \ref{table-final}.

It's worth mentioning that TRADES-AWP (TA) is much more robust than other models in our pool. Majority-vote ensemble accuracy is a factor of individual model robustnesss forming the ensemble, and their diverseness. As the rest of the models are comparatively quite less robust, adding them with TRADES-AWP to form an ensemble leads to comparatively less increase over the individual TRADES-AWP model (57.6\% to 59.84\%). However, just restricting to other models in the pool except TRADES-AWP leads to much higher increase. The highest base model robust accuracy is 54.86\% of MART, to 59.2\% of the best ensemble formed out of the remaining $6$ models, thereby an increase of $\sim$4.34\%.

\vspace{-10pt}\section{Conclusion}
We show voting-based ensemble can improve the robustness over a set of adversarially trained base models. Furthermore, we propose an  algorithm to automatically select an ensemble from a pool of models. Using our algorithm, we achieve an ensemble with state-of-the-art robust accuracy on CIFAR-10 dataset, and the performance is verified by both existing and several newly proposed strong attacks. 


{\small
\bibliographystyle{plainnat}
\bibliography{egbib}

\begin{thebibliography}{46}
\providecommand{\natexlab}[1]{#1}
\providecommand{\url}[1]{\texttt{#1}}
\expandafter\ifx\csname urlstyle\endcsname\relax
  \providecommand{\doi}[1]{doi: #1}\else
  \providecommand{\doi}{doi: \begingroup \urlstyle{rm}\Url}\fi

\bibitem[Alayrac et~al.(2019)Alayrac, Uesato, Huang, Fawzi, Stanforth, and
  Kohli]{alayrac2019labels}
Jean-Baptiste Alayrac, Jonathan Uesato, Po-Sen Huang, Alhussein Fawzi, Robert
  Stanforth, and Pushmeet Kohli.
\newblock Are labels required for improving adversarial robustness?
\newblock In \emph{Advances in Neural Information Processing Systems}, pages
  12214--12223, 2019.

\bibitem[Andriushchenko et~al.(2020)Andriushchenko, Croce, Flammarion, and
  Hein]{andriushchenko2020square}
Maksym Andriushchenko, Francesco Croce, Nicolas Flammarion, and Matthias Hein.
\newblock Square attack: a query-efficient black-box adversarial attack via
  random search, 2020.

\bibitem[Athalye et~al.(2018)Athalye, Carlini, and
  Wagner]{athalye2018obfuscated}
Anish Athalye, Nicholas Carlini, and David Wagner.
\newblock Obfuscated gradients give a false sense of security: Circumventing
  defenses to adversarial examples.
\newblock \emph{International Coference on International Conference on Machine
  Learning}, 2018.

\bibitem[Biggio et~al.(2013)Biggio, Corona, Maiorca, Nelson, {\v{S}}rndi{\'c},
  Laskov, Giacinto, and Roli]{biggio2013evasion}
Battista Biggio, Igino Corona, Davide Maiorca, Blaine Nelson, Nedim
  {\v{S}}rndi{\'c}, Pavel Laskov, Giorgio Giacinto, and Fabio Roli.
\newblock Evasion attacks against machine learning at test time.
\newblock In \emph{Joint European conference on machine learning and knowledge
  discovery in databases}, pages 387--402, 2013.

\bibitem[Brendel et~al.(2019)Brendel, Rauber, Kümmerer, Ustyuzhaninov, and
  Bethge]{brendel2019accurate}
Wieland Brendel, Jonas Rauber, Matthias Kümmerer, Ivan Ustyuzhaninov, and
  Matthias Bethge.
\newblock Accurate, reliable and fast robustness evaluation, 2019.

\bibitem[Carlini and Wagner(2017)]{Carlini_2017}
Nicholas Carlini and David Wagner.
\newblock Towards evaluating the robustness of neural networks.
\newblock \emph{2017 IEEE Symposium on Security and Privacy (SP)}, May 2017.
\newblock \doi{10.1109/sp.2017.49}.
\newblock URL \url{http://dx.doi.org/10.1109/SP.2017.49}.

\bibitem[Carmon et~al.(2019)Carmon, Raghunathan, Schmidt, Duchi, and
  Liang]{carmon2019unlabeled}
Yair Carmon, Aditi Raghunathan, Ludwig Schmidt, John~C Duchi, and Percy~S
  Liang.
\newblock Unlabeled data improves adversarial robustness.
\newblock In \emph{Advances in Neural Information Processing Systems}, pages
  11192--11203, 2019.

\bibitem[Chen and Gu(2020)]{chen2020rays}
Jinghui Chen and Quanquan Gu.
\newblock Rays: A ray searching method for hard-label adversarial attack, 2020.

\bibitem[Croce and Hein(2019)]{croce2019minimally}
Francesco Croce and Matthias Hein.
\newblock Minimally distorted adversarial examples with a fast adaptive
  boundary attack, 2019.

\bibitem[Croce and Hein(2020)]{croce2020reliable}
Francesco Croce and Matthias Hein.
\newblock Reliable evaluation of adversarial robustness with an ensemble of
  diverse parameter-free attacks, 2020.

\bibitem[Ding et~al.(2018)Ding, Sharma, Lui, and Huang]{ding2018max}
Gavin~Weiguang Ding, Yash Sharma, Kry Yik~Chau Lui, and Ruitong Huang.
\newblock Max-margin adversarial (mma) training: Direct input space margin
  maximization through adversarial training.
\newblock \emph{arXiv preprint arXiv:1812.02637}, 2018.

\bibitem[Ding et~al.(2019)Ding, Wang, and Jin]{ding2019advertorch}
Gavin~Weiguang Ding, Luyu Wang, and Xiaomeng Jin.
\newblock {AdverTorch} v0.1: An adversarial robustness toolbox based on
  pytorch.
\newblock \emph{arXiv preprint arXiv:1902.07623}, 2019.

\bibitem[Goodfellow et~al.(2015)Goodfellow, Shlens, and
  Szegedy]{goodfellow2014explaining}
Ian~J Goodfellow, Jonathon Shlens, and Christian Szegedy.
\newblock Explaining and harnessing adversarial examples.
\newblock \emph{International Conference on Learning Representations}, 2015.

\bibitem[He et~al.(2017)He, Wei, Chen, Carlini, and
  Song]{DBLP:journals/corr/HeWCCS17}
Warren He, James Wei, Xinyun Chen, Nicholas Carlini, and Dawn Song.
\newblock Adversarial example defenses: Ensembles of weak defenses are not
  strong.
\newblock \emph{CoRR}, abs/1706.04701, 2017.
\newblock URL \url{http://arxiv.org/abs/1706.04701}.

\bibitem[Huang et~al.(2020{\natexlab{a}})Huang, Zhang, and Zhang]{huang2020sat}
Lang Huang, Chao Zhang, and Hongyang Zhang.
\newblock Self-adaptive training: beyond empirical risk minimizatio.
\newblock \emph{arXiv preprint arXiv:2002.10319}, 2020{\natexlab{a}}.

\bibitem[Huang et~al.(2020{\natexlab{b}})Huang, Zhang, and
  Zhang]{huang2020selfadaptive}
Lang Huang, Chao Zhang, and Hongyang Zhang.
\newblock Self-adaptive training: beyond empirical risk minimization,
  2020{\natexlab{b}}.

\bibitem[Jang et~al.(2019)Jang, Zhao, Hong, and Lee]{Jang_2019_ICCV}
Yunseok Jang, Tianchen Zhao, Seunghoon Hong, and Honglak Lee.
\newblock Adversarial defense via learning to generate diverse attacks.
\newblock In \emph{Proceedings of the IEEE/CVF International Conference on
  Computer Vision (ICCV)}, October 2019.

\bibitem[Kurakin et~al.(2017)Kurakin, Goodfellow, and
  Bengio]{kurakin2016adversarial_ICLR}
Alexey Kurakin, Ian Goodfellow, and Samy Bengio.
\newblock Adversarial machine learning at scale.
\newblock \emph{International Conference on Learning Representations}, 2017.

\bibitem[Liu et~al.(2018)Liu, Cheng, Zhang, and Hsieh]{liu2018towards}
Xuanqing Liu, Minhao Cheng, Huan Zhang, and Cho-Jui Hsieh.
\newblock Towards robust neural networks via random self-ensemble.
\newblock In \emph{Proceedings of the European Conference on Computer Vision
  (ECCV)}, pages 369--385, 2018.

\bibitem[Madry et~al.(2017)Madry, Makelov, Schmidt, Tsipras, and
  Vladu]{aleks2017deep}
Aleksander Madry, Aleksandar Makelov, Ludwig Schmidt, Dimitris Tsipras, and
  Adrian Vladu.
\newblock Towards deep learning models resistant to adversarial attacks, 2017.

\bibitem[Madry et~al.(2018)Madry, Makelov, Schmidt, Tsipras, and
  Vladu]{madry2017towards}
Aleksander Madry, Aleksandar Makelov, Ludwig Schmidt, Dimitris Tsipras, and
  Adrian Vladu.
\newblock Towards deep learning models resistant to adversarial attacks.
\newblock \emph{International Conference on Learning Representations}, 2018.

\bibitem[Moosavi-Dezfooli et~al.(2016)Moosavi-Dezfooli, Fawzi, and
  Frossard]{moosavidezfooli2016deepfool}
Seyed-Mohsen Moosavi-Dezfooli, Alhussein Fawzi, and Pascal Frossard.
\newblock Deepfool: a simple and accurate method to fool deep neural networks,
  2016.

\bibitem[Mustafa et~al.(2019)Mustafa, Khan, Hayat, Goecke, Shen, and
  Shao]{Mustafa_2019_ICCV}
Aamir Mustafa, Salman Khan, Munawar Hayat, Roland Goecke, Jianbing Shen, and
  Ling Shao.
\newblock Adversarial defense by restricting the hidden space of deep neural
  networks.
\newblock In \emph{Proceedings of the IEEE/CVF International Conference on
  Computer Vision (ICCV)}, October 2019.

\bibitem[Pang et~al.(2019)Pang, Xu, Du, Chen, and
  Zhu]{DBLP:journals/corr/abs-1901-08846}
Tianyu Pang, Kun Xu, Chao Du, Ning Chen, and Jun Zhu.
\newblock Improving adversarial robustness via promoting ensemble diversity.
\newblock \emph{CoRR}, abs/1901.08846, 2019.
\newblock URL \url{http://arxiv.org/abs/1901.08846}.

\bibitem[Pang et~al.(2020)Pang, Yang, Dong, Xu, Su, and Zhu]{pang2020boosting}
Tianyu Pang, Xiao Yang, Yinpeng Dong, Kun Xu, Hang Su, and Jun Zhu.
\newblock Boosting adversarial training with hypersphere embedding, 2020.

\bibitem[Qiu et~al.(2019)Qiu, Leng, Guo, Chen, Li, Guo, and Zhu]{Qiu_2019_CVPR}
Yuxian Qiu, Jingwen Leng, Cong Guo, Quan Chen, Chao Li, Minyi Guo, and Yuhao
  Zhu.
\newblock Adversarial defense through network profiling based path extraction.
\newblock In \emph{Proceedings of the IEEE/CVF Conference on Computer Vision
  and Pattern Recognition (CVPR)}, June 2019.

\bibitem[Rahnama et~al.(2020)Rahnama, Nguyen, and Raff]{Rahnama_2020_CVPR}
Arash Rahnama, Andre~T. Nguyen, and Edward Raff.
\newblock Robust design of deep neural networks against adversarial attacks
  based on lyapunov theory.
\newblock In \emph{Proceedings of the IEEE/CVF Conference on Computer Vision
  and Pattern Recognition (CVPR)}, June 2020.

\bibitem[Rauber et~al.(2017)Rauber, Brendel, and Bethge]{rauber2017foolbox}
Jonas Rauber, Wieland Brendel, and Matthias Bethge.
\newblock Foolbox: A python toolbox to benchmark the robustness of machine
  learning models.
\newblock In \emph{Reliable Machine Learning in the Wild Workshop, 34th
  International Conference on Machine Learning}, 2017.
\newblock URL \url{http://arxiv.org/abs/1707.04131}.

\bibitem[Rauber et~al.(2020)Rauber, Zimmermann, Bethge, and
  Brendel]{rauber2017foolboxnative}
Jonas Rauber, Roland Zimmermann, Matthias Bethge, and Wieland Brendel.
\newblock Foolbox native: Fast adversarial attacks to benchmark the robustness
  of machine learning models in pytorch, tensorflow, and jax.
\newblock \emph{Journal of Open Source Software}, 5\penalty0 (53):\penalty0
  2607, 2020.
\newblock \doi{10.21105/joss.02607}.
\newblock URL \url{https://doi.org/10.21105/joss.02607}.

\bibitem[Schmidt et~al.(2018)Schmidt, Santurkar, Tsipras, Talwar, and
  Madry]{schmidt2018adversarially}
Ludwig Schmidt, Shibani Santurkar, Dimitris Tsipras, Kunal Talwar, and
  Aleksander Madry.
\newblock Adversarially robust generalization requires more data.
\newblock In \emph{Advances in Neural Information Processing Systems}, pages
  5014--5026, 2018.

\bibitem[Sen et~al.(2020)Sen, Ravindran, and Raghunathan]{sen2020empir}
Sanchari Sen, Balaraman Ravindran, and Anand Raghunathan.
\newblock Empir: Ensembles of mixed precision deep networks for increased
  robustness against adversarial attacks, 2020.

\bibitem[Sitawarin et~al.(2020)Sitawarin, Chakraborty, and
  Wagner]{sitawarin2020improving}
Chawin Sitawarin, Supriyo Chakraborty, and David Wagner.
\newblock Improving adversarial robustness through progressive hardening, 2020.

\bibitem[Strauss et~al.(2017)Strauss, Hanselmann, Junginger, and
  Ulmer]{strauss2017ensemble}
Thilo Strauss, Markus Hanselmann, Andrej Junginger, and Holger Ulmer.
\newblock Ensemble methods as a defense to adversarial perturbations against
  deep neural networks, 2017.

\bibitem[Szegedy et~al.(2014{\natexlab{a}})Szegedy, Zaremba, Sutskever, Bruna,
  Erhan, Goodfellow, and Fergus]{Pertub}
C.~Szegedy, W.~Zaremba, I.~Sutskever, J.~Bruna, D.~Erhan, I.~Goodfellow, and
  R.~Fergus.
\newblock Intriguing properties of neural networks.
\newblock \emph{International Conference on Learning Representations},
  2014{\natexlab{a}}.

\bibitem[Szegedy et~al.(2014{\natexlab{b}})Szegedy, Zaremba, Sutskever, Bruna,
  Erhan, Goodfellow, and Fergus]{szegedy2013intriguing}
Christian Szegedy, Wojciech Zaremba, Ilya Sutskever, Joan Bruna, Dumitru Erhan,
  Ian Goodfellow, and Rob Fergus.
\newblock Intriguing properties of neural networks.
\newblock \emph{International Conference on Learning Representations},
  2014{\natexlab{b}}.

\bibitem[Szegedy et~al.(2014{\natexlab{c}})Szegedy, Zaremba, Sutskever, Bruna,
  Erhan, Goodfellow, and Fergus]{szegedy2014intriguing}
Christian Szegedy, Wojciech Zaremba, Ilya Sutskever, Joan Bruna, Dumitru Erhan,
  Ian Goodfellow, and Rob Fergus.
\newblock Intriguing properties of neural networks, 2014{\natexlab{c}}.

\bibitem[Tramer et~al.(2020)Tramer, Carlini, Brendel, and
  Madry]{tramer2020adaptive}
Florian Tramer, Nicholas Carlini, Wieland Brendel, and Aleksander Madry.
\newblock On adaptive attacks to adversarial example defenses, 2020.

\bibitem[Tramèr et~al.(2017)Tramèr, Kurakin, Papernot, Goodfellow, Boneh, and
  McDaniel]{tramr2017ensemble}
Florian Tramèr, Alexey Kurakin, Nicolas Papernot, Ian Goodfellow, Dan Boneh,
  and Patrick McDaniel.
\newblock Ensemble adversarial training: Attacks and defenses, 2017.

\bibitem[Verma and Swami(2019)]{NEURIPS2019_cd61a580}
Gunjan Verma and Ananthram Swami.
\newblock Error correcting output codes improve probability estimation and
  adversarial robustness of deep neural networks.
\newblock In H.~Wallach, H.~Larochelle, A.~Beygelzimer, F.~d\textquotesingle
  Alch\'{e}-Buc, E.~Fox, and R.~Garnett, editors, \emph{Advances in Neural
  Information Processing Systems}, volume~32, pages 8646--8656. Curran
  Associates, Inc., 2019.
\newblock URL
  \url{https://proceedings.neurips.cc/paper/2019/file/cd61a580392a70389e27b0bc2b439f49-Paper.pdf}.

\bibitem[Wang and Zhang(2019)]{Wang_2019_ICCV}
Jianyu Wang and Haichao Zhang.
\newblock Bilateral adversarial training: Towards fast training of more robust
  models against adversarial attacks.
\newblock In \emph{Proceedings of the IEEE/CVF International Conference on
  Computer Vision (ICCV)}, October 2019.

\bibitem[Wang et~al.(2020)Wang, Zou, Yi, Bailey, Ma, and Gu]{Wang2020Improving}
Yisen Wang, Difan Zou, Jinfeng Yi, James Bailey, Xingjun Ma, and Quanquan Gu.
\newblock Improving adversarial robustness requires revisiting misclassified
  examples.
\newblock In \emph{International Conference on Learning Representations}, 2020.
\newblock URL \url{https://openreview.net/forum?id=rklOg6EFwS}.

\bibitem[Wu et~al.(2020)Wu, Xia, and Wang]{wu2020adversarial}
Dongxian Wu, Shu-Tao Xia, and Yisen Wang.
\newblock Adversarial weight perturbation helps robust generalization.
\newblock In \emph{NeurIPS}, 2020.

\bibitem[Xie et~al.(2019)Xie, Wu, Maaten, Yuille, and He]{Xie_2019_CVPR}
Cihang Xie, Yuxin Wu, Laurens van~der Maaten, Alan~L. Yuille, and Kaiming He.
\newblock Feature denoising for improving adversarial robustness.
\newblock In \emph{Proceedings of the IEEE/CVF Conference on Computer Vision
  and Pattern Recognition (CVPR)}, June 2019.

\bibitem[Xiong and Hsieh(2020)]{xiong2020improved}
Yuanhao Xiong and Cho-Jui Hsieh.
\newblock Improved adversarial training via learned optimizer, 2020.

\bibitem[Zhang et~al.(2019)Zhang, Yu, Jiao, Xing, Ghaoui, and
  Jordan]{zhang2019theoretically}
Hongyang Zhang, Yaodong Yu, Jiantao Jiao, Eric~P Xing, Laurent~El Ghaoui, and
  Michael~I Jordan.
\newblock Theoretically principled trade-off between robustness and accuracy.
\newblock \emph{International Conference on Machine Learning}, 2019.

\bibitem[Zheng et~al.(2020)Zheng, Zhang, Gu, Lee, and Prakash]{Zheng_2020_CVPR}
Haizhong Zheng, Ziqi Zhang, Juncheng Gu, Honglak Lee, and Atul Prakash.
\newblock Efficient adversarial training with transferable adversarial
  examples.
\newblock In \emph{Proceedings of the IEEE/CVF Conference on Computer Vision
  and Pattern Recognition (CVPR)}, June 2020.

\end{thebibliography}
}
\newpage
\appendix

\section{Appendix}
\subsection{Algorithm for choosing models to ensemble} \label{appendix-select-model-ens}

We present {\sc Choose\_Ensemble} in Algorithm \ref{alg:select-model-ens-0} for selecting models for majority-vote ensemble mentioned in Section \ref{select-model-ens-0}. We denote by $A(\cdot)$ a generic white-box attacks which takes a model and a point $x_i$ and generates an adversarial example. We don't include other details like perturbation bound and other hyperparameters for brevity.\\

\begin{figure}[htbp]
    \centering				
    \label{fig:heuristic}
    \begin{tabular}{c}
    
        \subfloat[accuracy on WA-Autoattack]{\includegraphics[width=0.4\textwidth]{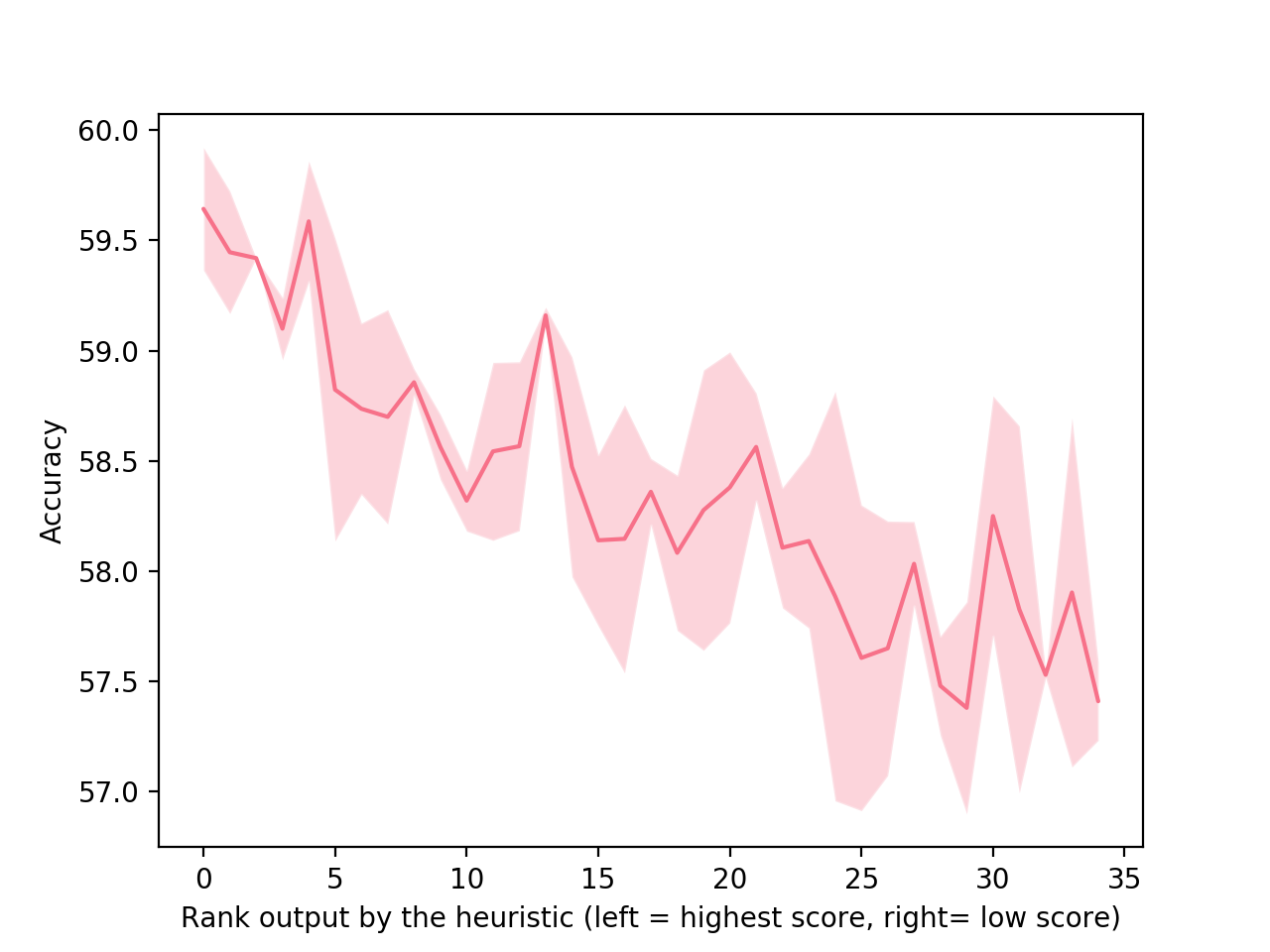}}\\
		\subfloat[accuracy on Strong WA-FAB attack]{\includegraphics[width=0.4\textwidth]{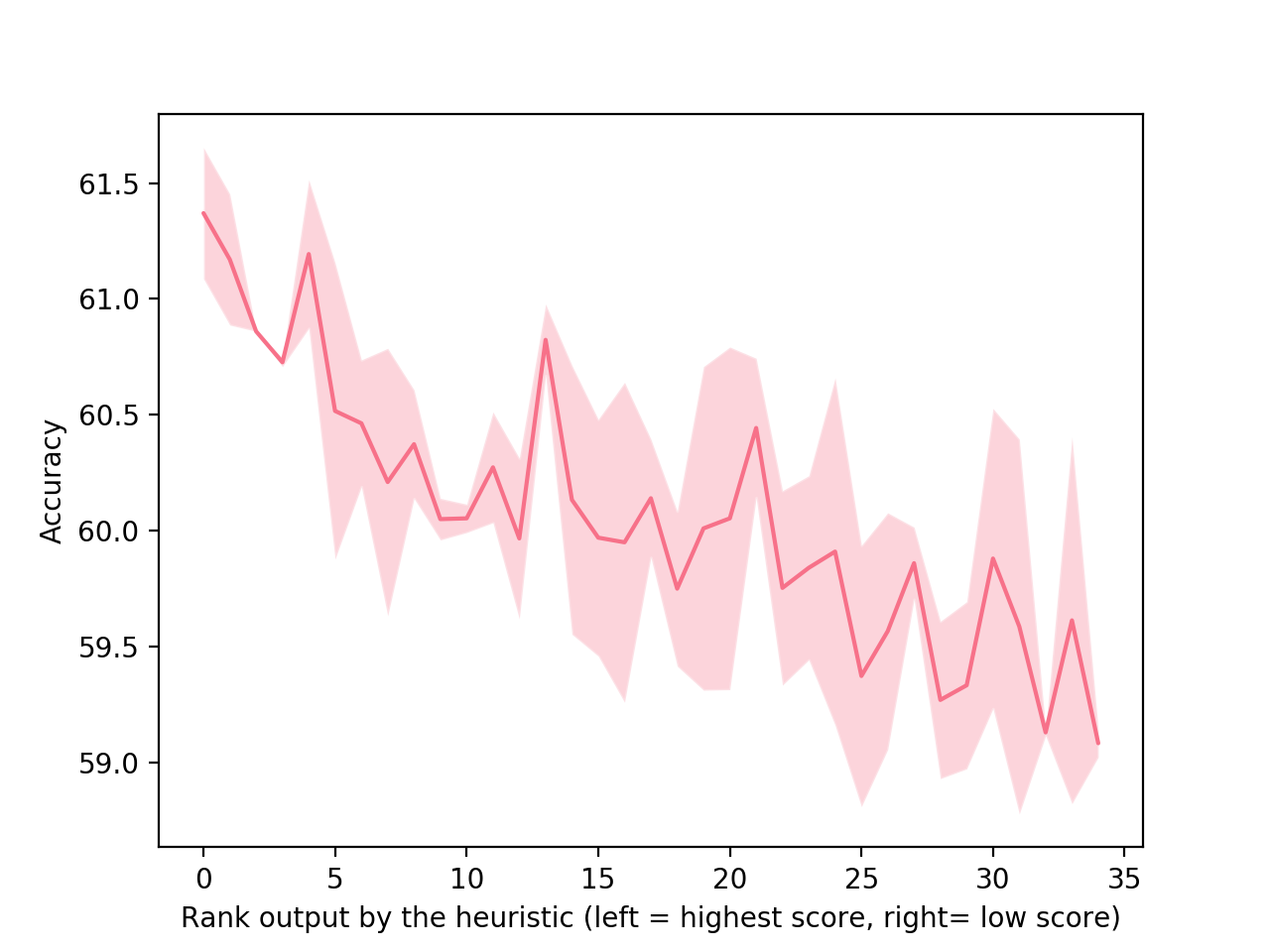}}
	\end{tabular}
	    \caption{Average accuracy vs rank (correspondingly score order) of the heuristic proposed. The shaded area represents the error region}

\end{figure}  

\begin{algorithm}[htbp]
\caption{{\sc Choose\_Ensemble}$(A(\cdot),X,Y,M,k)$}
\label{alg:select-model-ens-0}
\begin{algorithmic}[1]
\State \textbf{Input:} Attack function $A(\cdot)$, test points $(X,Y)$, $M=\{f_0,\cdots,f_{n-1}\}$ base models, size of ensemble to form $k$
\State set $G=\{g_i(\cdot) \ | \ i\in \left[{n\choose k}\right]\}$ \Comment{Form a set of all possible $\left[{n\choose k}\right]\}$ ensembles}
\For{$x_i$ in $X$}
\State set $\mathbf{x}_i' \gets$ A($f_{i\%n}, x$) \Comment{Attack $(i\%n)^{th} model$}
\EndFor
\For{$g_i$ in G}
\State set $sc(g_i) \gets \sum_{j=1}^{n}\mathbbm{1}_{g_i(\mathbf{x}_j')==y_j}$
\EndFor
\State \textbf{return} $\{sc(g_i) \ | \ i\in[{n\choose k}] \}$
\end{algorithmic}
\end{algorithm}

\noindent We use this algorithm in Section \ref{select-model-ens} on a pool of $7$ models to form $3$ model ensembles, hence ${7\choose3} = 35$ ensembles. We plot the output ranks (in decreasing scores) with the actual accuracy of the ensemble in Figure 1. We take three runs of the algorithm over random $\sim 4500$ points and plot the mean accuracy for each rank. The shaded area represents the error region (mean\_accuracy - std\_dev\_of\_accuracy to mean\_accuracy + std\_dev\_of\_accuracy). We observe that our algorithm performs reasonably well with the best ensemble always in top $3$ scored, and an overall consistent trend of accuracy-vs-rank. Recall that the average (over $3$ runs) Kendell's $\tau$ statistic of predicted score by the algorithm and the actual accuracy of the ensembles is $+0.53$ and $+0.503$ for WA-autoattack and WA-FAB attack respectively.

\subsection{Algorithm for weakest-attacked attack technique} \label{appendix-WA}
\noindent We present {\sc Weakest\_Attacked}$(\cdot)$ in Algorithm \ref{alg:weakest-attacked}. This is the weakest-attacked attack strategy mentioned in Section \ref{WA}. 

\begin{algorithm}[htbp]
\caption{{\sc Weakest\_Attacked}$(A(\cdot),g(\cdot),lr,ss,s,\mathcal{B}, \mathbf{x}, y)$}
\label{alg:weakest-attacked}
\begin{algorithmic}[1]
\State \textbf{Input:} Attack function $A(\cdot)$, data point and true label $(\mathbf{x},y)$, learning rate $lr$, step size $ss$, number of steps $s$, allowed perturbation set $\mathcal{B}$, and the majority-vote ensemble $g(\cdot)$.
\State set $\mathbf{x\_adv} \gets \mathbf{x}$
\For{i in s}
\If{$g(\mathbf{x\_adv})\neq y$}
\State Break
\EndIf
\State min\_prob $\gets 1$
\For{$model$ in $g$}
\State set $p \gets Softmax(model(x\_adv))$
\If{$\arg\max_{j}p_j == y$ and $p_y \leq $min\_prob}
\State min\_prob = $p_y$
\State model\_to\_attack $\gets model$
\EndIf
\EndFor
\State $\mathbf{x\_adv} \gets A(\text{model\_to\_attack}, lr, ss, s=1, \mathbf{x\_adv}, y)$
\State $x\_adv \gets \Pi_\mathcal{B}(x\_adv-x)+x$ \Comment{Project the perturbation on allowed perturbation set}
\EndFor
\State \textbf{return} $\mathbf{x\_adv}$
\end{algorithmic}
\end{algorithm}

\end{document}